\documentclass[a4paper]{article}

\usepackage{times}
\usepackage{latexsym}

\usepackage[english]{babel}
\usepackage[utf8x]{inputenc}
\usepackage{microtype}
\usepackage[colorinlistoftodos]{todonotes}
\usepackage[margin=1.2in]{geometry}

\usepackage{hyperref}
\usepackage{url}
\usepackage{mathtools}

\usepackage{amsmath, amssymb}
\usepackage{ntheorem}

\usepackage{graphicx}
\usepackage{caption}
\usepackage{subcaption}
\usepackage{multirow}
\usepackage{color}

\usepackage{soul}
\usepackage{times}
\usepackage{latexsym}
\usepackage{algorithm}
\usepackage{algorithmic}
\usepackage{graphicx}
\usepackage{inconsolata}
\usepackage{caption}
\usepackage{subcaption}
\usepackage{amsmath}
\usepackage{booktabs}

\usepackage{natbib}

\title{SwitchCIT: Switching for Continual Instruction Tuning}

\author{
  \textbf{Xinbo Wu}$^{1,2}$, \textbf{Max Hartman}$^{2}$\textbf{,}  \textbf{Vidhata Arjun Jayaraman}$^{2,3}$\textbf{,} \textbf{Lav R. Varshney }$^{1,2}$\\
  $^1$Coordinated Science Laboratory \\
  $^2$Department of Electrical and Computer Engineering \\
  $^3$Department of Mathematics \\
  University of Illinois Urbana-Champaign\\ 
  \texttt{\{xinbowu2, maxh3, vidhata2, varshney\}@illinois.edu}
  }
  
\date{}
\begin{document}
\maketitle
\begin{abstract}
Large language models (LLMs) and multimodal models (MMs) have exhibited impressive capabilities in various domains, particularly in general language understanding and visual reasoning. However, these models, trained on massive data, may not be finely optimized for specific tasks triggered by instructions. Continual instruction tuning is crucial to adapt a large model to evolving tasks and domains, ensuring their effectiveness and relevance across a wide range of applications. In the context of continual instruction tuning, where models are sequentially trained on different tasks, catastrophic forgetting can occur, leading to performance degradation on previously learned tasks. This work addresses the catastrophic forgetting in continual instruction learning through a switching mechanism for routing computations to parameter-efficient tuned models. We demonstrate the effectiveness of our method through experiments on continual instruction tuning of different natural language generation tasks and vision-language tasks. We also showcase the advantages of our proposed method in terms of efficiency, scalability, portability, and privacy preservation.
\end{abstract}

\section{Introduction}
Large language models (LLMs) and multimodal models (MMs) have demonstrated remarkable capabilities across numerous domains, as highlighted by \citet{GPT-4} and \citet{bubeck2023sparks}. However, whereas both LLMs and MMs pre-trained on extensive data excel in general language understanding and multimodal tasks, they may not be optimized for every specific task of interest prompted by instructions. Therefore, there is a need for continual instruction learning to adapt LLMs and MMs to evolving tasks and domains. Indeed, continual instruction learning is essential for LLMs such as GPT \citep{Radford2019LanguageMA} and MMs such as LLaVA \citep{liu2024visual, liu2024improved} to maintain their effectiveness and relevance in handling a wide range of tasks and domains. 

Such models are trained on vast amounts of data and fine-tuned for specific applications, often by learning tasks sequentially \citep{luo2023empirical}, i.e.\ learning on datasets pertaining to one task all at once, before moving on to the next task. The challenge lies in their ability to continually learn and adapt as they encounter new tasks and information. However, in continual instruction learning scenarios, where models are sequentially trained on different tasks or datasets, catastrophic forgetting occurs when the model’s parameters are updated to accommodate new information, leading to degradation or complete loss of performance on previously learned tasks. 

A typical way to balance new learning with the retention of previously acquired capabilities in LLMs is through replaying old data. However, with the rapid iterations of LLMs for diverse and complex use cases, retaining old data becomes exceptionally challenging. Moreover, continually tuning a model with a large number of parameters is highly costly in terms of both computation and memory usage. Parameter-efficient fine-tuning (PEFT) such as low-rank adaptation (LoRA)~\citep{lora} provides an option of lightweight with portable parameters, which could be paired with a model to perform specific tasks. Therefore, in this work, we focus on alleviating catastrophic forgetting during continual instruction tuning, particularly with minimal data retention and its interplay with PEFT. 

We propose a novel continual instruction tuning method, SwitchCIT, that alleviates forgetting of previously seen tasks by introducing a switch network to identify a task for a user query, leveraging the clustering phenomenon of task-specific instruction vectors~\citep{clustering_llm}. For each new task, we fine-tune the task performance by including extra parameters created by PEFT methods such as LoRA (a self-expansion process) in addition to a pre-trained large base model making the method more practical. 

Our method represents a distinct paradigm compared to the current mainstream machine learning (ML) system design, where a single large model simultaneously or continually learns to perform as many tasks as possible. This difference may encourage a rethinking or reformulation of existing ML system designs.

Catastrophic forgetting in neural networks is related to the palimpsest phenomenon that new memories rapidly overwrite old ones ~\citep{zenke2024theories}. SwitchCIT may be considered as a way to avoid the need to overwrite the old memories of previously learned tasks by introducing extra parameters for a new task. Moreover, there exists a line of methods inspired by synaptic consolidation in brains that reduces the learning rate on specific weights based on their importance to previously encountered tasks. Our method has the advantage of enabling the full tuning of all weights to adapt to a specific task without restricting any particular set of weights. 

A recent study \citep{anonymous2024spurious} attributes forgetting from fine-tuning a large model to the learned bias of its alignment toward a specific task but without fundamentally altering the model's transferable knowledge across tasks. Our proposed paradigm aligns well with this perspective: we fine-tune separate models for distinct tasks to prevent potential forgetting of prior task alignments and avoid conflicts among alignments for different tasks. While a limitation of this paradigm is its ineffectiveness in leveraging transferable knowledge across tasks, this concern is alleviated in the era of large models; we expect most essential transferable knowledge to be acquired during the pre-training stage, reducing the need for further obtaining it during continual fine-tuning.

We summarize our contributions as follows:
\begin{itemize}
    \item We propose a novel continual instruction-tuning approach to alleviate catastrophic forgetting by using a switch network for task routing to different specialized models tuned via PEFT.
    \item We conduct experiments for instruction-tuning on five continual natural language generation tasks and two multimodal vision-language tasks, demonstrating the effectiveness of our method compared to several baselines.
    \item We demonstrate additional advantages of our method including the efficiency, scalability, portability, and privacy preservation of our proposed method.
\end{itemize}

\section{Related Work}

\textbf{Continual Learning and Catastrophic Forgetting.} The study of continual learning focuses on developing algorithms that learn from a continuous stream of data, enabling a model to acquire new knowledge while retaining previously learned information without catastrophic forgetting \citep{wang2024comprehensive}. Catastrophic forgetting happens when LLMs forget previously learned information as new tasks are learned \citep{luo2023empirical}. \citet{anonymous2024amuro} provides an insightful study empirically showing that pre-trained LLMs may forget domain knowledge and tasks that were not included in the fine-tuning process, while supervised fine-tuning offers substantial benefits to the models. To counter this effect, here we use a switch network to classify tasks and route computations from their instructions. By doing so, we can fine-tune task performance by including extra parameters created by PEFT methods such as LoRA for each task. Continual learning methods are often classified into the following categories: replay-based methods \citep{chaudhry2019tiny,chaudhry2018efficient}, regularization-based methods \citep{hu2019overcoming,buzzega2020dark}, and architecture-based methods \citep{veniatefficient,yan2021dynamically}. Catastrophic forgetting arises when model parameters are updated to account for new data, causing degradation or a complete loss of performance on previously learned tasks. Catastrophic forgetting is not specific to LLMs. Other neural networks also experience this phenomenon, leading to methods such as Elastic Weight Consolidation \citep{kirkpatrick2017overcoming}. Previous research to resolve this problem has scaled the number of parameters in the model \citep{wang2024self, yoon2018lifelong}. It has been demonstrated that these solutions work in theory but suffer from over-reliance on scaling LLM parameters. Works such as \citet{hu2019overcoming} avoid this by splitting parameters into two sets: one for tasks learned and one to dynamically be generated. 

\textbf{Understanding Transformers.} Prior studies have offered insightful understandings of Transformer models with focuses on the internal representations \citep{clustering_llm, wu2023meta, nanda2023emergent} and attention mechanisms \citep{sun2021effective,olsson2022context}. Inspired by \citet{clustering_llm}, we design a novel method for continual instruction tuning for LLMs via switching instead of concentrating on understanding of Transformers. 

\textbf{Instruction Tuning.} Instruction tuning is the process of tuning a model from specific instructions or prompts that will guide the model toward behaving in the desired fashion. 
A major issue with LLMs has been the mismatch between the training objective of the LLM and users’ objectives. Instruction tuning has been developed, in part, to address this issue. This method of training aims to align language models with human intent \citep{ouyang2022training, stiennon2020learning, zhang2023instruction}. Moreover, instruction tuning has proven to be highly effective in multimodal settings.~\citep{dai2023instructblip, liu2024visual, liu2024improved}. We concentrate our work on the specific case of continual instruction tuning across different tasks, which presents unique challenges such as catastrophic forgetting. Models trained by instruction tuning have been shown to have better performance on unseen tasks than without \citep{wei2021finetuned}. Instruction tuning, however, has its challenges of crafting high-quality instructions that can cover the desired behavior, and it seems to only capture surface-level patterns rather than truly comprehending the task 

\textbf{Parameter-Efficient Fine-Tuning.} PEFT addresses the challenge of needing enormous computing resources to fine-tune contemporary LLMs. PEFT reduces the number of fine-tuning parameters and memory usage while still achieving similar results as full fine-tuning \citep{xu2023parameterefficient}. One particularly popular PEFT method is LoRA, which freezes the model weights of the pre-trained model and injects trainable rank decomposition matrices into each layer of the Transformer architecture, allowing training on a small number of additional parameters rather than on the original pre-trained model \citep{hu2019overcoming}. Here, we use LoRA to create extra parameters for fine-tuning tasks not yet seen by the LLM. Since the original proposal of LoRA, there have been many derivatives that aim to resolve certain limitations of the original LoRA method, \citep{valipour2023dylora, zhang2023adalora, dettmers2023qlora, huang2024lorahub}. Notwithstanding, LoRA still seems to be the most widely used PEFT method.

\textbf{Mixture of Experts.} Mixture of experts (MoE) models integrate multiple sub-models, or experts, to address different parts of the input space ~\citep{jacobs1991adaptive,du2022glam,zoph2022st}. Though the MoE and ours share some similarities, SwitchCIT uses different models to handle different parts of a task space represented by instructions, rather than an arbitrary input space as in MoE models. Also, we separate the learning processes for model selection and the models themselves, whereas MoE models learn both simultaneously. SwitchCIT can self-expand its parameters to adapt to new tasks, whereas MoE models typically do not.

\section{Preliminaries}
Consider a model parameterized by $\theta$, which is trained to optimize a loss function $\mathcal{L}$ while incorporating human-provided instructions $\{I_t\}_{t=1}^T$ over a sequence of tasks $T$.

The objective of continual instruction tuning is to update the model parameters $\theta$ in response to new data received during the learning process. The formulation involves iteratively adapting the model's behavior to align with the evolving set of instruction-based data. 

Let's assume there are $N_{t}$ data examples for a task $t$: $\{(I_{t},x_{t,n},y_{t,n})\}_{n=1}^{N_{t}}$, where an instruction $I_{t}$, an input $x_{t,n}$ and an output $y_{t,n}$ form a tuple. The continual instruction tuning framework can be defined as follows:

\[
\min_{\theta} \sum_{t=1}^T \sum_{n=1}^{N_{t}} \mathcal{L}_{t}(f_\theta([I_{t};x_{t,n}]), y_{t,n}) + \lambda R(\theta)
\]

where:
\begin{itemize}
    \item $[I_{t};x_{t,n}]$ is a concatenation of an instruction and an input, where the input $x_{t,n}$ may belong to a different modality, such as vision and is optional.  
    \item $f_\theta([I_{t};x_{t,n}])$ denotes the output of the model $f$ parameterized by $\theta$ for the task $t$ with an input $x_{t,n}$.
    \item $y_{t,n}$ represents the target labels or outputs associated with task $t$.
    \item $\mathcal{L}$ is the standard loss function measuring the discrepancy between model predictions and true labels for task $t$.
    \item $R(\theta)$ is an optional regularization term that penalizes changes to the model parameters $\theta$, helping to mitigate catastrophic forgetting.
    \item $\lambda$ is a hyperparameter that controls the trade-off between optimizing for task-specific performance and preserving knowledge from previous tasks.
\end{itemize}

In this formulation, the model parameters $\theta$ are updated sequentially across tasks $t=1, \dots, T$, where each iteration involves minimizing a task-specific loss. The continual instruction tuning objective aims to update the model's parameters $\theta$ incrementally across tasks, optimizing performance on new tasks while maintaining competence on previously learned tasks. This formulation embodies the principle of lifelong learning, enabling the model to adapt and improve over time without sacrificing past knowledge. Many classical methods such as ~\citet{kirkpatrick2017overcoming} rely on the regularization term $R(\theta)$ to alleviate catastrophic forgetting. However, our proposed method approaches the catastrophic forgetting problem differently, without relying on regularization, as will be discussed in the following section.

\section{Method}

\begin{figure*}
\centering
\includegraphics[width=0.6\linewidth]{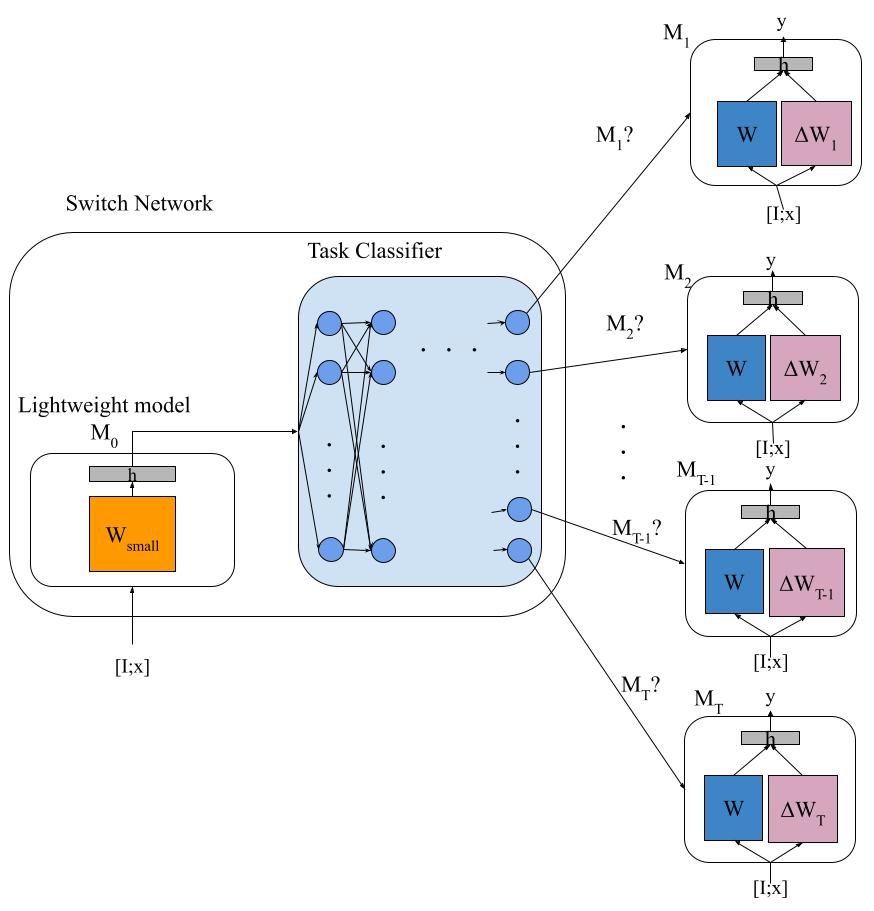}
\caption{The inference procedure of SwitchCIT. The instruction is first fed into a switch network consisting of a lightweight LLM and a task classifier. The last token representation from the final layer of the LLM is used as the input to the task classifier. This switch network classifies a given task and then routes computation to the associated set of parameters.}
\label{fig:method}
\end{figure*}
 
We illustrate SwitchCIT in Figure~\ref{fig:method}. At inference time, a tuple $[I; x]$ is given, where $I$ is an instruction and $x$ is an optional input. Based on the instruction $I$, a switch network routes the computation to a model trained explicitly for the predicted task such that the performance of both previously learned and newly learned tasks is mostly retained. More specifically, the switch network identifies tasks via a multi-class classification from their instructions for the routing by using instruction features extracted by a lightweight LLM, $W_{small}$. We use the last token representation of an instruction from the final layer of $W_{small}$ as the features. This design is inspired by the fact that vector representations of instructions belonging to the same task are clustered together within the hidden representation space, with the task-specific clustering phenomenon becoming more pronounced in later layers \citep{clustering_llm}. Note that effective clustering implies good separability of task representations. 

A selected model relies on a concatenation of the instruction and the input, $[I; x]$ to anticipate an output $y$ via an internal representation $h$ produced by a base model $W$ and its task-specific weight $\Delta{W}$. For brevity, we omit details about the computations of $h$ and of reaching $y$ from $h$, which involves a causal decoding process; see \citet{vaswani2017attention, lora} for more details. Therefore, the switch network allows tasks to be handled by models dedicated to them. Models tailored to different tasks will not interfere with one another, which consequently alleviates catastrophic forgetting of previously learned task. Here, both $x$ and $y$ could be considered as textual sequences in the context of language generation. All models $M_1$--$M_T$ are instruction-tuned for different tasks ($1$ to $T$) by introducing extra parameters $\Delta{W}$ through a PEFT method such as LoRA. The switch network may be easily implemented as a multi-layer perceptron (MLP) model with an instruction feature encoder such as an LLM.

\section{Experiments on Language Tasks}\label{sec:explanguage}
In this section, we briefly overview the model implementation and datasets. Experimental setups are further detailed in Appendices~\ref{appx:implementation} and \ref{appx:tasks}. 

We use BLOOMZ 1.1B and BLOOMZ 7.1B ~\citep{bloomz} as two choices of our base LLMs and LoRA~\citep{lora} to learn task-specific and portable parameters. A switch network is implemented using a two-layer MLP network. We use OPT-125M~\citep{zhang2022opt}, a lightweight LLM with only 125 million parameters to extract features for the switch network. Implementation and training details are in Appendix~\ref{appx:implementation}. 

We consider the benchmark introduced by~\citet{scialom2022fine} and five continual instruction tasks for natural language generation selected by~\citet{luo2023empirical} to differ from the training and evaluation tasks of BLOOMZ. These tasks are: 
\begin{itemize}
    \item Text Simplification (Simp), \citep{jiang-etal-2020-neural} and \citep{alva-manchego-etal-2020-asset}, paraphrasing a given text into simpler language; 
    \item Empathetic Dialogue Generation (Emdg) \citep{rashkin2019empathetic}, generating a dialogue response that offers a reason within a specified emotional context.
    \item Inquisitive Question Generation (InqQG) \citep{fan2019eli5}, create questions that prompt long-form answers. 
    \item Explanation Generation (Exp), according to \citet{NEURIPS2018_4c7a167b}, this involves generating natural language explanations for provided premises, hypotheses, or labels. 
    \item Headline Generation with Constraint (HGen) \citep{scialom2022fine}, which focuses on producing headlines that include specific keywords, positioned as per the constraints specified. We evaluate a model's performance on each task using metrics outlined by~\citet{scialom2022fine}.
\end{itemize}
Specifically, we use SARI for Simp; BERTSore (BS) for Emdg, InqQA, and Exp; Rouge-1 (R1) for HGen. We use the BERTScore version based on DeBERTa-MNLI.

Following \citet{luo2023empirical}, we train a model based on the order of instruction tasks: Simp → Emdg → InqQG → Exp → HGen. We use the specific prompts designed by \citet{scialom2022fine} and train on 100,000 data samples, following previous works~\citep{luo2023empirical,scialom2022fine}. We feed an instruction possible with an optional input to the switch network for determining task identity since in practice, an instruction and an input may not be easy to separate.

We compare our method to a direct supervised fine-tuning approach and a continual learning method with a rehearsal mechanism following \citet{scialom2022fine}. We consider the rehearsal method because it is the only method in the literature that is developed in an experimental setting closest to our continual instruction tuning setting and a representative of continual learning methods based on data replaying. Both our method and the other methods employ the same PEFT using LoRA unless specified otherwise. 

To ensure a fair comparison, we use the same amount of data for rehearsal as we do for training our switch network. To investigate the regime of low data retaining, we reserve only 0.01\% of the training data for each task to train the switch network, which is significantly less than the amount of replay data such as the 1\% used by traditional continual learning methods for rehearsal. We also evaluate methods vi using 1\% of the training data, based on which the rehearsal method was originally developed. Details of the instructions, prompts, and training are in Appendix~\ref{appx:tasks}. We evaluate these methods on test splits of various task datasets as detailed in Appendix~\ref{appx:tasks}.

\begin{table*}
\begin{center}
\begin{tabular}{lllll}
\hline
\multicolumn{1}{c}{\bf Retention Rate} & \multicolumn{1}{c}{\bf Emdg}  & \multicolumn{1}{c}{\bf InqQG} & \multicolumn{1}{c}{\bf Exp} & \multicolumn{1}{c}{\bf HGen}
\\ \hline 

0.01 \% & 100.0 \% & 96.3 \% & 97.6 \% & 97.1 \%\\
1.0 \% & 100.0 \% & 99.9 \% & 99.9 \% & 99.6 \%\\
\hline
\end{tabular}
\caption{Progressive performance of task classification by our switch networks measured by accuracy. A continual learning stage is denoted by its task name. Note that there is no switch network for the first learned task. }
\label{tab:switch_network}
\end{center}
\end{table*}

\subsection{Switch Network}
Table~\ref{tab:switch_network} presents the progressive performance of task classification by our switch networks trained under different conditions: a low data retention rate-0.01\% and 1\% comparable to the replay data used by the rehearsal method of \citet{scialom2022fine}. Note that after learning each task, we retrain a very lightweight task classifier to accommodate the newly learned task. It is evident that switch networks of different settings achieve very high classification accuracy at every learning stage, even when using a lightweight LLM like the OPT-125M for feature extraction. The performance is scaled up by using more training data. Performance does not obviously degrade when including more tasks, demonstrating good robustness. Notice that competitive classification performance is reached even with 100X less data, which may be explained by good separability of different tasks due to task clustering shown by \citet{clustering_llm}.

\subsection{Continual Instruction Tuning}

The results presented in Table~\ref{tab:cl_1} are derived from experiments conducted with the base LLM, BLOOMZ-1.1B, retaining 1\% training data. As shown in Table~\ref{tab:cl_1}, the performance of the model continually tuned with Parameter-Efficient Fine-Tuning (PEFT) degrades significantly on previously learned tasks compared to the initial model, as expected. However, our proposed method, SwitchCIT, demonstrates substantial improvements over the initial model across all tasks. This indicates that SwitchCIT effectively mitigates forgetting and reaps the benefits of continual instruction tuning. Notably, SwitchCIT outperforms the PEFT model on the HGen task, which is the last task tuned, suggesting that our approach may effectively reduce negative interference from prior tasks. Furthermore, SwitchCIT exceeds the performance of the rehearsal method that uses the same retention data rate (1\%) as studied by \citet{scialom2022fine}. This comparison highlights that our method more effectively utilizes the retained data than rehearsing previous data.

\begin{table*}
\begin{center}
\begin{tabular}{llllll}
\hline
\multicolumn{1}{c}{\bf Method}  & \multicolumn{1}{c}{\bf Simp}  & \multicolumn{1}{c}{\bf Emdg}  & \multicolumn{1}{c}{\bf InqQG} & \multicolumn{1}{c}{\bf Exp} & \multicolumn{1}{c}{\bf HGen}
\\ \hline 
Initial & 37.7 & 0.483 & 0.457 & 0.515 & 0.269\\
PEFT & 34.9 & 0.265 & 0.454 & 0.369 & 0.356\\
Rehearsal & 48.4 & 0.533 & 0.589 & 0.685 & 0.357\\
SwitchCIT & \textbf{49.0} & \textbf{0.546} & \textbf{0.593} & \textbf{0.712} & \textbf{0.359}\\
\hline
\end{tabular}
\caption{The final performance of various methods on different tasks in the continual instruction tuning. Tasks are presented in the learning order. We also list the performance of the original LLM as "Initial".}
\label{tab:cl_1}
\end{center}
\end{table*}

We present studies on using base LLMs of varying scales and different data retention rates in Table~\ref{tab:cl_2}. We compare SwitchCIT (BLOOMZ-1.1B and 0.01\% data retention) relying on PEFT to a much larger fully supervised fine-tuned approach based on the BLOOMZ-7.1b model \citep{luo2023empirical,bloomz}, which is approximately seven times larger than our base LLM. Surprisingly, our method still achieves better performance on most tasks, underscoring its efficiency and effectiveness. We also include results using the larger base LLM, BLOOMZ-7.1B, which achieves the best performance with 1\% data retention, demonstrating the scalability of our method. In contrast to the rehearsal method, our approach experiences notably less performance degradation when retaining 100 times less data from 1\% to 0.01\%. This performance gap across most tasks also remains minimal for our SwitchCIT, when using the larger base LLM, BLOOMZ-7.1B. We hypothesize that due to our high-performing switch network, our method is able to specifically select a tailored sub-model for a task, yielding impressive performance. This also suggests that SwitchCIT requires much less data retention compared to the rehearsal method, highlighting the practicality and efficiency of our approach.

\begin{table*}
\begin{center}
\begin{tabular}{llllllll}
\hline
\multicolumn{1}{c}{\bf Method}  & \multicolumn{1}{c}{\bf Base LLM}  & \multicolumn{1}{c}{\bf Retention Rate}  & \multicolumn{1}{c}{\bf Simp}  & \multicolumn{1}{c}{\bf Emdg}  & \multicolumn{1}{c}{\bf InqQG} & \multicolumn{1}{c}{\bf Exp} & \multicolumn{1}{c}{\bf HGen}
\\ \hline 
Full-SFT & BLOOMZ-7.1B &  & 47.2 & 0.533 & 0.597 & 0.687 & 0.329\\
Rehearsal & BLOOMZ-1.1B & 0.01\% & 36.8 & 0.458 & 0.484 & 0.587 &
0.359\\
Rehearsal & BLOOMZ-1.1B & 1\% & 48.4 & 0.533 & 0.589 & 0.685 & 0.357\\
SwitchCIT & BLOOMZ-1.1B &  0.01\% & 49.0 & 0.545 & 0.559 & 0.712 & 0.355\\
SwitchCIT & BLOOMZ-1.1B &  1\% & 49.0 & 0.546 & 0.593 & 0.712 & 0.359\\
SwitchCIT & BLOOMZ-7.1B &  0.01\% & \textbf{49.5} & 0.561 & 0.577 & \textbf{0.726} & 0.414\\
SwitchCIT & BLOOMZ-7.1B &  1\% & \textbf{49.5} & \textbf{0.562} & \textbf{0.615} & \textbf{0.726} & \textbf{0.418}\\
\hline
\end{tabular}
\caption{The final performance of various methods in different settings on different tasks in the continual instruction tuning. Tasks are presented in the learning order. Full-SFT refers to the fully supervised fine-tuning method without using the PEFT.}
\label{tab:cl_2}
\end{center}
\end{table*}

To evaluate the success of the continual instruction tuning over a sequence of \(N\) tasks, we use the concept of Upper Bound (UB) by following ~\citet{scialom2022fine}. The UB represents the highest performance that the model can achieve when fine-tuned exclusively on a specific task, \(T_n\). The model is considered to be successful if it maintains a performance close to the UB while learning new tasks. The normalized results, referred to as Relative Gain for a given task \(T_n\), are calculated as: 
\begin{equation}
\begin{aligned}
   RG_{T_n} = \frac{s_{T_n}}{UB_{T_n}}
\label{eqt:RG}
\end{aligned}
\end{equation}

where $s_{T_n}$ is the performance of a continually tuned model on task $T_n$. A high relative gain indicates effective retention of performance on a specific task and a value of 1 indicates that the model performs similarly to the UB for a task. 

\begin{figure*}
\centering
\begin{subfigure}{0.33\textwidth}
  \centering
  \includegraphics[width=1.0\linewidth]{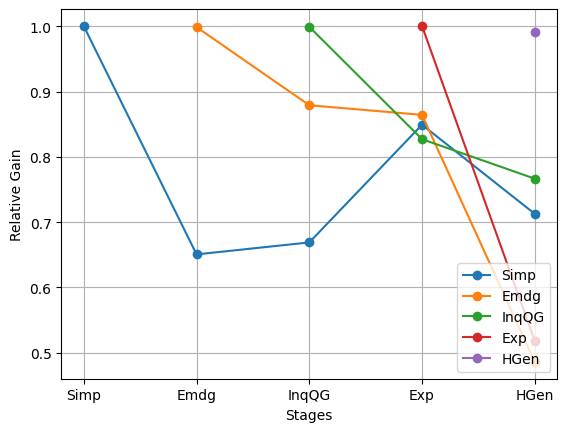}
  \caption{PEFT}
  \label{fig:forgeting_sft}
\end{subfigure}%
\begin{subfigure}{0.33\textwidth}
  \centering
  \includegraphics[width=1.0\linewidth]{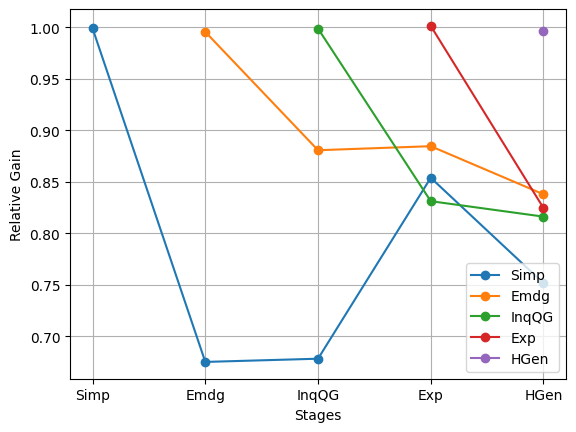}
  \caption{Rehearsal}
  \label{fig:forgeting_replay}
\end{subfigure}%
\begin{subfigure}{0.33\textwidth}
  \centering
  \includegraphics[width=1.0\linewidth]{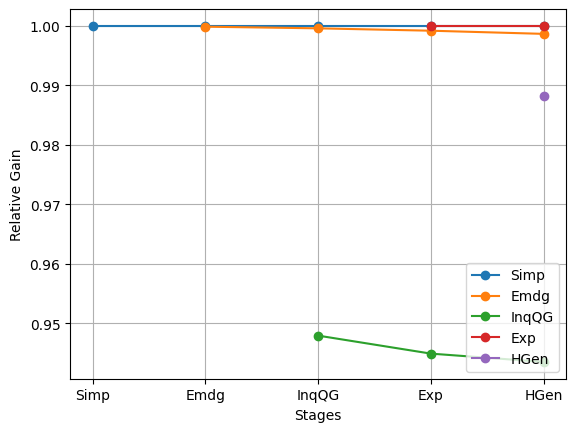}
  \caption{SwitchCIT}
  \label{fig:forgeting_ours}
\end{subfigure}%
\caption{Progressive relative gain of various methods. The horizontal axis presents different learning stages labeled by their respective task names, whereas the vertical axis shows the relative gain. Task performances are shown once the task is learned across different stages.}
\label{fig:forgeting}
\end{figure*}

From Figure~\ref{fig:forgeting}, notice that SwitchCIT experiences minimal catastrophic forgetting compared to other approaches, as shown by almost perfect retention of performance for various tasks. The less perfect retention on the InqQG task is due to imperfect task identification by the switch network. In contrast to our method, the rehearsal method retains only some performance on previous tasks. Along with PEFT, they both exhibit noticeable levels of catastrophic forgetting as they continue to learn additional tasks. 

\section{Experiments on Multimodal Tasks}
Multimodal tasks are problems that involve processing and understanding information from multiple data modalities (e.g., text, images, audio, video, structured data). These tasks leverage the complementary strengths of each modality to enhance performance and enable capabilities that are otherwise impossible with a single modality. 

Models for handling vision-language tasks have been widely studied. Instruction tuning is a crucial step for vision-language models (VLMs) because it allows them to generalize better, align more closely with user intentions, and handle a wide variety of tasks by following explicit natural language instructions. However, current VLMs are typically trained in a multitask setting, and continual instruction tuning for VLMs remains largely unexplored. This is becoming increasingly demanding as vision-language tasks people expect a system to handle grow more complex and diverse.

We focused our experiments on a recent and widely adopted vision-language model (VLM), LLaVA v1.5 ~\citep{liu2024improved}, which has demonstrated impressive performance on vision-language tasks in a multitask setting with high training efficiency. Its training pipeline includes a pre-training stage followed by an instruction-tuning stage. In this work, we decompose its original multitask instruction tuning stage into a continual instruction tuning of individual tasks.

We study two distinct tasks: Visual Reasoning (VR) and OCR-based Visual Question Answering (OCR-VQA), as they are sufficiently different and have well-established benchmarks. For the VR task, we utilize the GQA dataset ~\citep{hudson2019gqa} for both training and evaluation. For OCR-VQA, we perform training on the OCR-VQA dataset ~\citep{mishraICDAR19} and evaluation on the scene text-centric VQA data from the OCRBench benchmark ~\citep{liu2023hidden}. Task performance is reported in terms of accuracy. We adopt a continual learning order of GQA → OCR-VQA.

Our approach follows the parameter-efficient instruction tuning strategy with LoRA proposed by ~\citet{liu2024improved}. Specifically, we use Vicuna 7B v1.5 ~\citep{liu2024improved} as the base LLM and employ the same switch network setup described in Section ~\ref{sec:explanguage}. For comparison, we include a rehearsal-based baseline that relies on 0.01\% data replay.

\begin{figure*}
\centering
\begin{subfigure}{0.33\textwidth}
  \centering
  \includegraphics[width=1.0\linewidth]{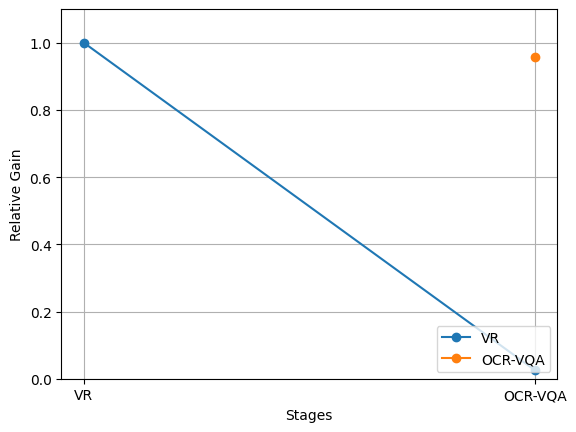}
  \caption{Rehearsal}
  \label{fig:forgetting_replay_vlm}
\end{subfigure}%
\begin{subfigure}{0.33\textwidth}
  \centering
  \includegraphics[width=1.0\linewidth]{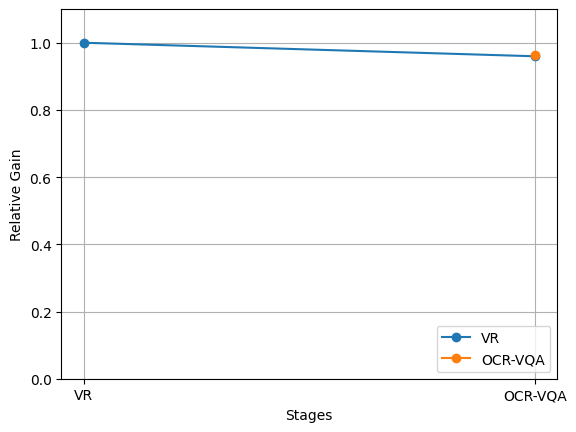}
  \caption{SwitchCIT}
  \label{fig:forgeting_ours_vlm}
\end{subfigure}%
\caption{Progressive relative gain of various methods. The horizontal axis presents different learning stages labeled by their respective task names, whereas the vertical axis shows the relative gain. Task performances are shown once the task is learned across different stages.}
\label{fig:forgeting_vlm}
\end{figure*}

Table ~\ref{tab:cl_vl} and Figure ~\ref{fig:forgeting_ours_vlm} present the comparison between our method and the rehearsal-based baseline. The results reveal that even with data replay, the VLM exhibits a significantly worse forgetting issue for a previously learned task compared to its LLM counterparts. We hypothesize that this is because VLMs often receive similar visual inputs across tasks, making it struggle to distinguish among them. In contrast, our SwitchCIT method effectively mitigates the forgetting issue, as evidenced by the high relative gain shown in Figure ~\ref{fig:forgeting_ours_vlm}. This demonstrates not only the effectiveness of our approach but also its robustness across different application domains.

\begin{table*}
\begin{center}
\begin{tabular}{lll}
\hline
\multicolumn{1}{c}{\bf Method}  & \multicolumn{1}{c}{\bf VR}  & \multicolumn{1}{c}{\bf OCR-VQA} 
\\ \hline 
Rehearsal & 0.0159 & 0.545 \\
SwitchCIT & \textbf{0.595} & \textbf{0.550}\\
\hline
\end{tabular}
\caption{The final performance of various methods on different vision-language tasks in the continual instruction tuning. Tasks are presented in the learning order.}
\label{tab:cl_vl}
\end{center}
\end{table*}

\section{Additional Advantages}
We will discuss additional advantages beyond the superior performance of our method in this section.

\subsection{Efficiency}
Many existing works overcome catastrophic forgetting by imposing constraints on the existing parameters of a model, so their model sizes will not change. SwitchCIT introduces new parameters for each additional task. For example, when using BLOOMZ 1.1B as the base LLM, these additional parameters account for only 0.878\% of the total parameters. Considering five continual tasks, the additional parameters amount to just 4.39\% in exchange for minimal catastrophic forgetting, demonstrating their lightweight and practical feasibility. Note that only the additional parameters specific to a single task are loaded during inference. We anticipate further improvements in these numbers as parameter-efficient methods continue to advance. 

\subsection{Scalability}
The size of a multitask learner usually needs to scale with the number of tasks to maintain high performance across all tasks. However, this scaling can lead to increased response latency as the number of tasks grows. In contrast, our method enables the creation of a constant-size model tailored to new tasks while maintaining strong overall performance. This ensures that latency is minimally affected as the number of tasks increases; the lightweight switch network may scale slowly with the number of tasks, but the resulting latency increase remains minimal. This is particularly crucial as we expect systems to become increasingly versatile in the future, yet minimizing latency is essential for scenarios where timely responses are critical, such as medical diagnosis and financial trading.

\subsection{Portability}
Separating the development of the switch network from the instruction-tuned models greatly enhances SwitchCIT's portability. For instance, to improve task identification by our switch network using more data (from 0.01\% to 1.0\%, as shown in Table ~\ref{tab:switch_network}), we only need to retrain the switch network and plug it in the existing instruction-tuned models. Conversely, we can also use existing switch networks for better instruction-tuned models as shown in Table ~\ref{tab:cl_2}, where we leverage the same switch network for models with larger base LLMs such as BLOOMZ 7.1B.

\subsection{Privacy Preservation}
Privacy is a crucial consideration as the data used to train, test, and deploy models often includes sensitive, personal, or proprietary information. Ensuring privacy not only safeguards individual rights and organizational interests but also promotes ethical, legal, and trust-related aspects of our intelligent systems. Our parameter-efficient task-specific models can be trained separately and distributively, offering a robust privacy-preserving capability. For instance, sensitive data, such as medical records, can remain localized for model training, while the trained model can still be utilized during inference. This privacy-preserving feature could potentially forge collaborations among different industries, where one party provides the technology, and another contributes application-critical but sensitive data.

\section{Conclusion and Discussion}
We proposed a novel continual instruction-tuning approach to alleviate catastrophic forgetting by using a switch network to identify tasks and then route computations to parameter-efficient tuned models. Experiments conducted on five instruction-based continual natural language generation tasks and two continual vision-language tasks demonstrate the effectiveness of our method compared to several baselines. We also analyze and show the efficiency, scalability, portability, and privacy preservation of our proposed method. Note that our method is broadly applicable to the standard pre-training and fine-tuning framework for large models by treating these two stages as two sequentially learned tasks and addressing the issue of potential forgetting accordingly. However, how to view pre-training and fine-tuning as two distinct tasks is to be carefully defined based on specific use cases, ensuring that the switch network can reliably identify the appropriate task for a given user query.

\section{Limitations}
Because of computational constraints, we could only test our method on models of limited scales. However, according to our design, our high-performing switch network is independent of the base models and can be paired with larger-scale models that offer superior performance. The remarkable performance of our switch network showcases the effectiveness of our method. It is worth noting that our task classifier in the switch network is incredibly lightweight (154K parameters) and requires minimal data (0.01\% of the training data), making it highly practical and easy to integrate.

\bibliography{main}

\begin{thebibliography}{}

\bibitem[Alva-Manchego et~al., 2020]{alva-manchego-etal-2020-asset}
Alva-Manchego, F., Martin, L., Bordes, A., Scarton, C., Sagot, B., and Specia, L. (2020).
\newblock {ASSET}: {A} dataset for tuning and evaluation of sentence simplification models with multiple rewriting transformations.
\newblock In Jurafsky, D., Chai, J., Schluter, N., and Tetreault, J., editors, {\em Proceedings of the 58th Annual Meeting of the Association for Computational Linguistics}.

\bibitem[Anonymous, 2024a]{anonymous2024amuro}
Anonymous (2024a).
\newblock Amuro and char: Analyzing the relationship between pre-training and fine-tuning of large language models.
\newblock In {\em Submitted to ACL Rolling Review - June 2024}.
\newblock under review.

\bibitem[Anonymous, 2024b]{anonymous2024spurious}
Anonymous (2024b).
\newblock Spurious forgetting in continual learning of language models.
\newblock In {\em Submitted to The Thirteenth International Conference on Learning Representations}.
\newblock under review.

\bibitem[Bubeck et~al., 2023]{bubeck2023sparks}
Bubeck, S., Chandrasekaran, V., Eldan, R., Gehrke, J., Horvitz, E., Kamar, E., Lee, P., Lee, Y.~T., Li, Y., Lundberg, S., Nori, H., Palangi, H., Ribeiro, M.~T., and Zhang, Y. (2023).
\newblock Sparks of artificial general intelligence: Early experiments with {GPT-4}.
\newblock {arXiv:2303.12712 [cs.CL].}

\bibitem[Buzzega et~al., 2020]{buzzega2020dark}
Buzzega, P., Boschini, M., Porrello, A., Abati, D., and Calderara, S. (2020).
\newblock Dark experience for general continual learning: a strong, simple baseline.
\newblock {\em Advances in neural information processing systems}, 33:15920--15930.

\bibitem[Camburu et~al., 2018]{NEURIPS2018_4c7a167b}
Camburu, O.-M., Rockt\"{a}schel, T., Lukasiewicz, T., and Blunsom, P. (2018).
\newblock {e-SNLI}: Natural language inference with natural language explanations.
\newblock In {\em Advances in Neural Information Processing Systems}, volume~31.

\bibitem[Chaudhry et~al., 2018]{chaudhry2018efficient}
Chaudhry, A., Ranzato, M., Rohrbach, M., and Elhoseiny, M. (2018).
\newblock Efficient lifelong learning with a-gem.
\newblock In {\em International Conference on Learning Representations}.

\bibitem[Chaudhry et~al., 2019]{chaudhry2019tiny}
Chaudhry, A., Rohrbach, M., Elhoseiny, M., Ajanthan, T., Dokania, P.~K., Torr, P.~H., and Ranzato, M. (2019).
\newblock On tiny episodic memories in continual learning.
\newblock {\em arXiv preprint arXiv:1902.10486}.

\bibitem[Dai et~al., 2023]{dai2023instructblip}
Dai, W., Li, J., Li, D., Tiong, A., Zhao, J., Wang, W., Li, B., Fung, P., and Hoi, S. (2023).
\newblock Instruct{BLIP}: Towards general-purpose vision-language models with instruction tuning.
\newblock In {\em Thirty-seventh Conference on Neural Information Processing Systems}.

\bibitem[Dettmers et~al., 2023]{dettmers2023qlora}
Dettmers, T., Pagnoni, A., Holtzman, A., and Zettlemoyer, L. (2023).
\newblock {QLoRA}: Efficient finetuning of quantized {LLMs}.
\newblock In {\em Advances in Neural Information Processing Systems}, volume~36, pages 10088--10115.

\bibitem[Du et~al., 2022]{du2022glam}
Du, N., Huang, Y., Dai, A.~M., Tong, S., Lepikhin, D., Xu, Y., Krikun, M., Zhou, Y., Yu, A.~W., Firat, O., et~al. (2022).
\newblock {GLaM}: Efficient scaling of language models with mixture-of-experts.
\newblock In {\em Proceedings of the 39th International Conference on Machine Learning}, pages 5547--5569.

\bibitem[Fan et~al., 2019]{fan2019eli5}
Fan, A., Jernite, Y., Perez, E., Grangier, D., Weston, J., and Auli, M. (2019).
\newblock {ELI5}: Long form question answering.
\newblock In {\em Proceedings of the 57th Annual Meeting of the Association for Computational Linguistics}, pages 3558--3567.

\bibitem[Hu et~al., 2022]{lora}
Hu, E.~J., Shen, Y., Wallis, P., Allen-Zhu, Z., Li, Y., Wang, S., Wang, L., and Chen, W. (2022).
\newblock {LoRA}: Low-rank adaptation of large language models.
\newblock In {\em Proceedings of the 10th International Conference on Learning Representations (ICLR)}.

\bibitem[Hu et~al., 2019]{hu2019overcoming}
Hu, W., Lin, Z., Liu, B., Tao, C., Tao, Z.~T., Zhao, D., Ma, J., and Yan, R. (2019).
\newblock Overcoming catastrophic forgetting for continual learning via model adaptation.
\newblock In {\em Proceedings of the 7th International Conference on Learning Representations (ICLR)}.

\bibitem[Huang et~al., 2023]{huang2024lorahub}
Huang, C., Liu, Q., Lin, B.~Y., Du, C., Pang, T., and Lin, M. (2023).
\newblock {LoraHub}: Efficient cross-task generalization via dynamic {LoRA} composition.
\newblock {arXiv:2307.13269 [cs.CL].}

\bibitem[Hudson and Manning, 2019]{hudson2019gqa}
Hudson, D.~A. and Manning, C.~D. (2019).
\newblock Gqa: A new dataset for real-world visual reasoning and compositional question answering.
\newblock In {\em Proceedings of the IEEE/CVF conference on computer vision and pattern recognition}, pages 6700--6709.

\bibitem[Jacobs et~al., 1991]{jacobs1991adaptive}
Jacobs, R.~A., Jordan, M.~I., Nowlan, S.~J., and Hinton, G.~E. (1991).
\newblock Adaptive mixtures of local experts.
\newblock {\em Neural Computation}, 3(1):79--87.

\bibitem[Jiang et~al., 2020]{jiang-etal-2020-neural}
Jiang, C., Maddela, M., Lan, W., Zhong, Y., and Xu, W. (2020).
\newblock Neural {CRF} model for sentence alignment in text simplification.
\newblock In {\em Proceedings of the 58th Annual Meeting of the Association for Computational Linguistics}, pages 7943--7960.

\bibitem[Kirkpatrick et~al., 2017]{kirkpatrick2017overcoming}
Kirkpatrick, J., Pascanu, R., Rabinowitz, N., Veness, J., Desjardins, G., Rusu, A.~A., Milan, K., Quan, J., Ramalho, T., Grabska-Barwinska, A., et~al. (2017).
\newblock Overcoming catastrophic forgetting in neural networks.
\newblock {\em Proceedings of the National Academy of Sciences}, 114(13):3521--3526.

\bibitem[Liu et~al., 2024a]{liu2024improved}
Liu, H., Li, C., Li, Y., and Lee, Y.~J. (2024a).
\newblock Improved baselines with visual instruction tuning.
\newblock In {\em Proceedings of the IEEE/CVF Conference on Computer Vision and Pattern Recognition}, pages 26296--26306.

\bibitem[Liu et~al., 2024b]{liu2024visual}
Liu, H., Li, C., Wu, Q., and Lee, Y.~J. (2024b).
\newblock Visual instruction tuning.
\newblock {\em Advances in neural information processing systems}, 36.

\bibitem[Liu et~al., 2023]{liu2023hidden}
Liu, Y., Li, Z., Yang, B., Li, C., Yin, X., Liu, C.-l., Jin, L., and Bai, X. (2023).
\newblock On the hidden mystery of ocr in large multimodal models.
\newblock {\em arXiv preprint arXiv:2305.07895}.

\bibitem[Luo et~al., 2023]{luo2023empirical}
Luo, Y., Yang, Z., Meng, F., Li, Y., Zhou, J., and Zhang, Y. (2023).
\newblock An empirical study of catastrophic forgetting in large language models during continual fine-tuning.
\newblock {arXiv:2308.08747 [cs.CL].}

\bibitem[Mishra et~al., 2019]{mishraICDAR19}
Mishra, A., Shekhar, S., Singh, A.~K., and Chakraborty, A. (2019).
\newblock Ocr-vqa: Visual question answering by reading text in images.
\newblock In {\em ICDAR}.

\bibitem[Muennighoff et~al., 2023]{bloomz}
Muennighoff, N., Wang, T., Sutawika, L., Roberts, A., Biderman, S., Scao, T.~L., Bari, M.~S., Shen, S., Yong, Z.-X., Schoelkopf, H., et~al. (2023).
\newblock Crosslingual generalization through multitask finetuning.
\newblock In {\em Proceedings of the 61st Annual Meeting of the Association for Computational Linguistics}, pages 15991--16111.

\bibitem[Nair and Hinton, 2010]{ReLU}
Nair, V. and Hinton, G.~E. (2010).
\newblock Rectified linear units improve restricted {B}oltzmann machines.
\newblock {\em Proceedings of the 27th International Conference on Machine Learning}, pages 807--814.

\bibitem[Nanda et~al., 2023]{nanda2023emergent}
Nanda, N., Lee, A., and Wattenberg, M. (2023).
\newblock Emergent linear representations in world models of self-supervised sequence models.
\newblock {\em arXiv preprint arXiv:2309.00941}.

\bibitem[Olsson et~al., 2022]{olsson2022context}
Olsson, C., Elhage, N., Nanda, N., Joseph, N., DasSarma, N., Henighan, T., Mann, B., Askell, A., Bai, Y., Chen, A., et~al. (2022).
\newblock In-context learning and induction heads.
\newblock {\em arXiv preprint arXiv:2209.11895}.

\bibitem[{OpenAI}, 2023]{GPT-4}
{OpenAI} (2023).
\newblock {GPT-4} technical report.
\newblock {arXiv:2304.01852 [cs.CL].}

\bibitem[Ouyang et~al., 2022]{ouyang2022training}
Ouyang, L., Wu, J., Jiang, X., Almeida, D., Wainwright, C., Mishkin, P., Zhang, C., Agarwal, S., Slama, K., Ray, A., et~al. (2022).
\newblock Training language models to follow instructions with human feedback.
\newblock In {\em Advances in Neural Information Processing Systems}, volume~35, pages 27730--27744.

\bibitem[Radford et~al., 2019]{Radford2019LanguageMA}
Radford, A., Wu, J., Child, R., Luan, D., Amodei, D., and Sutskever, I. (2019).
\newblock Language models are unsupervised multitask learners.

\bibitem[Rashkin et~al., 2019]{rashkin2019empathetic}
Rashkin, H., Smith, E.~M., Li, M., and Boureau, Y.-L. (2019).
\newblock Towards empathetic open-domain conversation models: a new benchmark and dataset.
\newblock In {\em Proceedings of the 57th Annual Meeting of the Association for Computational Linguistics}, pages 5370--5381.

\bibitem[Scialom et~al., 2022]{scialom2022fine}
Scialom, T., Chakrabarty, T., and Muresan, S. (2022).
\newblock Fine-tuned language models are continual learners.
\newblock In {\em Proceedings of the 2022 Conference on Empirical Methods in Natural Language Processing}, pages 6107--6122.

\bibitem[Stiennon et~al., 2020]{stiennon2020learning}
Stiennon, N., Ouyang, L., Wu, J., Ziegler, D., Lowe, R., Voss, C., Radford, A., Amodei, D., and Christiano, P.~F. (2020).
\newblock Learning to summarize with human feedback.
\newblock In {\em Advances in Neural Information Processing Systems}, volume~33, pages 3008--3021.

\bibitem[Sun and Marasovi{\'c}, 2021]{sun2021effective}
Sun, K. and Marasovi{\'c}, A. (2021).
\newblock Effective attention sheds light on interpretability.
\newblock In {\em Findings of the Association for Computational Linguistics: ACL-IJCNLP 2021}, pages 4126--4135.

\bibitem[Valipour et~al., 2023]{valipour2023dylora}
Valipour, M., Rezagholizadeh, M., Kobyzev, I., and Ghodsi, A. (2023).
\newblock {DyLoRA}: Parameter efficient tuning of pre-trained models using dynamic search-free low-rank adaptation.
\newblock In {\em Proceedings of the 17th Conference of the European Chapter of the Association for Computational Linguistics}, pages 3274--3287.

\bibitem[Vaswani et~al., 2017]{vaswani2017attention}
Vaswani, A., Shazeer, N., Parmar, N., Uszkoreit, J., Jones, L., Gomez, A.~N., Kaiser, {\L}., and Polosukhin, I. (2017).
\newblock Attention is all you need.
\newblock In {\em Advances in Neural Information Processing Systems}, volume~30.

\bibitem[Veniat et~al., 2021]{veniatefficient}
Veniat, T., Denoyer, L., and Ranzato, M. (2021).
\newblock Efficient continual learning with modular networks and task-driven priors.
\newblock In {\em International Conference on Learning Representations}.

\bibitem[Wang et~al., 2024a]{wang2024self}
Wang, H., Lu, H., Yao, L., and Gong, D. (2024a).
\newblock Self-expansion of pre-trained models with mixture of adapters for continual learning.
\newblock arXiv:2403.18886 [cs.LG].

\bibitem[Wang et~al., 2024b]{wang2024comprehensive}
Wang, L., Zhang, X., Su, H., and Zhu, J. (2024b).
\newblock A comprehensive survey of continual learning: Theory, method and application.
\newblock {\em IEEE Transactions on Pattern Analysis and Machine Intelligence}.
\newblock to appear.

\bibitem[Wei et~al., 2022]{wei2021finetuned}
Wei, J., Bosma, M., Zhao, V., Guu, K., Yu, A.~W., Lester, B., Du, N., Dai, A.~M., and Le, Q.~V. (2022).
\newblock Finetuned language models are zero-shot learners.
\newblock In {\em Proceedings of the 10th International Conference on Learning Representations (ICLR)}.

\bibitem[Wu and Varshney, 2023]{wu2023meta}
Wu, X. and Varshney, L.~R. (2023).
\newblock A meta-learning perspective on transformers for causal language modeling.
\newblock {\em arXiv preprint arXiv:2310.05884}.

\bibitem[Wu and Varshney, 2024]{clustering_llm}
Wu, X. and Varshney, L.~R. (2024).
\newblock Transformer-based causal language models perform clustering.
\newblock arXiv:2402.12151 [cs.CL].

\bibitem[Xu et~al., 2023]{xu2023parameterefficient}
Xu, L., Xie, H., Qin, S.-Z.~J., Tao, X., and Wang, F.~L. (2023).
\newblock Parameter-efficient fine-tuning methods for pretrained language models: A critical review and assessment.
\newblock {arXiv:2312.12148 [cs.CL].}

\bibitem[Yan et~al., 2021]{yan2021dynamically}
Yan, S., Xie, J., and He, X. (2021).
\newblock Der: Dynamically expandable representation for class incremental learning.
\newblock In {\em 2021 IEEE/CVF Conference on Computer Vision and Pattern Recognition (CVPR)}, pages 3013--3022. IEEE.

\bibitem[Yoon et~al., 2018]{yoon2018lifelong}
Yoon, J., Yang, E., Lee, J., and Hwang, S.~J. (2018).
\newblock Lifelong learning with dynamically expandable networks.
\newblock In {\em Proceedings of the 6th International Conference on Learning Representations (ICLR)}.

\bibitem[Zenke and Laborieux, 2024]{zenke2024theories}
Zenke, F. and Laborieux, A. (2024).
\newblock Theories of synaptic memory consolidation and intelligent plasticity for continual learning.
\newblock {arXiv:2405.16922 [q-bio.NC].}

\bibitem[Zhang et~al., 2023a]{zhang2023adalora}
Zhang, Q., Chen, M., Bukharin, A., He, P., Cheng, Y., Chen, W., and Zhao, T. (2023a).
\newblock Adaptive budget allocation for parameter-efficient fine-tuning.
\newblock In {\em Proceedings of the 11th International Conference on Learning Representations (ICLR)}.

\bibitem[Zhang et~al., 2023b]{zhang2023instruction}
Zhang, S., Dong, L., Li, X., Zhang, S., Sun, X., Wang, S., Li, J., Hu, R., Zhang, T., Wu, F., and Wang, G. (2023b).
\newblock Instruction tuning for large language models: A survey.
\newblock {arXiv:2308.10792 [cs.CL].}

\bibitem[Zhang et~al., 2022]{zhang2022opt}
Zhang, S., Roller, S., Goyal, N., Artetxe, M., Chen, M., Chen, S., Dewan, C., Diab, M., Li, X., Lin, X.~V., Mihaylov, T., Ott, M., Shleifer, S., Shuster, K., Simig, D., Koura, P.~S., Sridhar, A., Wang, T., and Zettlemoyer, L. (2022).
\newblock {OPT}: Open pre-trained transformer language models.
\newblock {2205.01068 [cs.CL].}

\bibitem[Zoph et~al., 2022]{zoph2022st}
Zoph, B., Bello, I., Kumar, S., Du, N., Huang, Y., Dean, J., Shazeer, N., and Fedus, W. (2022).
\newblock {ST-MoE}: Designing stable and transferable sparse expert models.
\newblock {2202.08906 [cs.CL].}

\end{thebibliography}
\bibliographystyle{apalike}

\newpage
\appendix

\section{Implementation Details}\label{appx:implementation}
We implement our switch network via OPT-125M ~\citep{zhang2022opt} as a feature extractor and its classifier using a two-layer MLP with ReLU activation function~\citep{ReLU}. We present hyperparameters related to the switch network in Table~\ref{table:switchnet_hyperparameters} and to continual learned models in Table~\ref{table:cl_hyperparameters}. We also present sizes of various used models in Table~\ref{table:model_size}. We perform the experiments on 2 X A100 Nvidia GPUs with 80 GB GPU memory.

\begin{table}
\centering
\begin{tabular}{lll}
\toprule
\textbf{Task}  & \textbf{Set} & \textbf{Size} \\
\hline
Simp & Training & 100,002\\
 & Testing & 4,000\\
Emdg & Training & 58,770\\
 & Testing & 8,396\\
InqQG & Training & 61,710\\
 & Testing & 1,681\\
Exp & Training & 100,002\\
 & Testing & 9,824\\
HGen & Training & 100,002\\
 & Testing & 1,951\\
\bottomrule
\end{tabular}
\caption{\label{table:data_statistics}
Data statistics of various tasks and their splits. 
}
\end{table}

\begin{table}
\centering
\begin{tabular}{ll}
\toprule
\textbf{Hyperparameter} & \textbf{Value} \\
\hline
Learning rate & 1E-3 \\
Number of epochs & 20 \\
Optimizer & AdamW \\
Scheduler & Constant \\
Number of layers & 2 \\
Hidden dimension & 200 \\
\bottomrule
\end{tabular}
\caption{\label{table:switchnet_hyperparameters}
Hyperparameters related to switch network and its training. 
}
\end{table}

\begin{table}
\centering
\begin{tabular}{ll}
\toprule
\textbf{Hyperparameter} & \textbf{Value} \\
\hline
Learning rate & 2E-5 \\
Number of epochs & 3 per task \\
Optimizer & AdamW \\
Scheduler & Constant \\
LoRA rank & 64 \\
LoRA Alpha & 16 \\
LoRA dropout & 0.05 \\
LoRA bias & None\\
\bottomrule
\end{tabular}
\caption{\label{table:cl_hyperparameters}
Hyperparameters related to models undergoing continual instruction tuning and their training. We adopt the training hyperparameters from \citet{luo2023empirical}.
}
\end{table}

\begin{table}

\centering
\begin{tabular}{ll}
\toprule
\textbf{Model} & \textbf{Parameter Count}\\
\hline
Switch Network &   154 thousand\\
OPT-125M &   125 million\\
BLOOMZ-1.1B  &  1.1 billion\\
BLOOMZ-7.1B  &  7.1 billion\\

\bottomrule
\end{tabular}
\caption{\label{table:model_size}Sizes of models used in this work in terms of parameter counts.
}
\end{table}

\section{Continual Instruction Tuning Tasks}\label{appx:tasks}
During instruction tuning, we follow~\citet{scialom2022fine} to start by adding a general prompt template at the beginning of the data: \textit{‘Below is an instruction that describes a task, paired with an input that provides further context. Write a response that appropriately completes the request...’}. This is then followed by a specific prompt for each task. We present statistics of different task datasets in Table~\ref{table:data_statistics}. In addition, we searched online and did not discover any information regarding the data containing information that names or uniquely identifies individual people or offensive content.

\section{License of Artifacts}
Our use of these artifacts is consistent with their intended use by following their respective protocols, see Table~\ref{table:license}. 
\begin{table}
\centering
\begin{tabular}{ll}
\toprule
\textbf{Name} & \textbf{License} \\
\hline
OPT-125M & OPT-125M License \\
BLOOMZ & The BigScience RAIL License \\
\bottomrule
\end{tabular}
\caption{\label{table:license}
License of artifacts used. 
}
\end{table}

\section{Potential Risks}
One potential risk associated with our approach is the possibility of making models preserve their bad behaviors. Future work will need to explore strategies for preventing this possibility.

\section{AI Assistance}
We used an LLM-based AI assistant (ChatGPT) to help in writing the paper.

\end{document}